\documentclass[
]{ceurart}

\usepackage{listings}

\usepackage{times}
\usepackage{latexsym}
\usepackage{comment}
\usepackage{algpseudocode}
\usepackage{algorithm}
\usepackage{arydshln}
\usepackage{float}

\begin{document}

\copyrightyear{2024}
\copyrightclause{Copyright for this paper by its authors. Use permitted under Creative Commons License Attribution 4.0 International (CC BY 4.0).}

\conference{Workshop on Psychology-informed Information Access Systems (PsyIAS), March 8, 2024, Mérida, Mexico}

\title{Can AI Models Appreciate Document Aesthetics? An Exploration of Legibility and Layout Quality in Relation to Prediction Confidence}


\author[1]{Hsiu-Wei Yang}[%
orcid=0009-0005-5630-1077,
email=leo.yang@thomsonreuters.com,
]
\address[1]{Thomson Reuters Labs, Toronto, Ontario, Canada}

\author[2]{Abhinav Agrawal}[%
orcid=0009-0006-2742-4569,
email=abhinav.agrawal@thomsonreuters.com,
]
\address[2]{Thomson Reuters Labs, Bangalore, Karnataka, India}

\author[3]{Pavlos Fragkogiannis}[%
orcid=0009-0001-0456-2743,
email=pavlos.fragkogiannis@thomsonreuters.com,
]
\address[3]{Thomson Reuters Labs, London, UK}

\author[2]{Shubham Nitin Mulay}[%
orcid=0009-0005-2519-5494,
email=shubham.nitinmulay@thomsonreuters.com,
]
\fnmark[1]

\fntext[1]{Work done during internship at Thomson Reuters Labs in 2023.}

\begin{abstract}
  A well-designed document communicates not only through its words but also through its visual eloquence. Authors utilize aesthetic elements such as colors, fonts, graphics, and layouts to shape the perception of information. Thoughtful document design, informed by psychological insights, enhances both the visual appeal and the comprehension of the content.
  While state-of-the-art document AI models demonstrate the benefits of incorporating layout and image data, it remains unclear whether the nuances of document aesthetics are effectively captured. 
  To bridge the gap between human cognition and AI interpretation of aesthetic elements, we formulated hypotheses concerning AI behavior in document understanding tasks, specifically anchored in document design principles. 
  With a focus on legibility and layout quality, we tested four aspects of aesthetic effects: noise, font-size contrast, alignment, and complexity, on model confidence using correlational analysis.
  The results and observations highlight the value of model analysis rooted in document design theories. 
  Our work serves as a trailhead for further studies and we advocate for continued research in this topic to deepen our understanding of how AI interprets document aesthetics.
\end{abstract}


\maketitle

\section{Introduction}
Reading documents involves complex cognitive functions such as perception, attention, memory, and comprehension. 
The aesthetic presentation of a document,  
such as font style, size, color, graphics, and page layout, 
can significantly impact how readers interact with it, influencing their understanding, retention, and overall engagement.
A well-crafted arrangement not only elevates the visual appeal of the document but also promotes the effectiveness of communication.  
For instance, an author can use typographical cues, such as underlining, bolding, and color, to guide readers in constructing perceptual rules for information seeking \cite{smith1983comprehension}, thereby improving content recall \cite{glynn1979control}.
In order to achieve desired communicative effects, the design industry has conducted extensive research on human reading behaviors, from which various design principles based on psychology, e.g., \emph{alignment} and \emph{contrast}, as well as evaluation criteria, e.g., \emph{legibility} and \emph{layout quality}, are established \cite{felker1980document,schriver1997dynamics,lentz2004functional, waller2011makes}.

From a machine learning standpoint, it is desirable for models to understand the semantic implications of aesthetics in high-quality documents, and to be resilient to subpar design.
In light of this, modern \emph{Document AI} \cite{cui2021document} systems lean towards multimodal Transformer architectures that integrate three sources of input or modalities, namely text, layout, and image \cite{huang2022layoutlmv3,  appalaraju2021docformer, powalski2021going}. Besides textual data (i.e., literal tokens), these models also take as inputs content element positions (i.e., coordinates of their bounding boxes) and document images (i.e., pixels) to learn layout and image representations, respectively. 
Empirical evidence highlights their efficiency in various benchmark tasks \cite{jaume2019funsd, huang2019icdar2019, zhong2019publaynet, pfitzmann2022doclaynet, mathew2021docvqa}. 
However, while the ablation studies affirm each modality's contribution to accuracy \cite{huang2022layoutlmv3, Gemelli_2023, appalaraju2021docformer, zhang2021entity, luo2023geolayoutlm, peng2022ernielayout}, without specifically addressing aesthetics in the model design, it is difficult to explain the predictions from these ``black boxes''.
Inspired by substantial work on interpreting text-only pre-trained models \cite{kovaleva2019revealing, jin2020bert, brunet2019understanding, belinkov2022probing}, we believe that a deeper understanding of these multimodal models can advance the field, prompting our research question:

\begin{itemize}
  \item \textbf{RQ:} \emph{Which aesthetic elements are perceived by the multimodal document AI systems when making predictions?} 
\end{itemize}

To address this, we laid the groundwork to bridge the current research gap. 
We studied literature on design principles and the impact of document aesthetics, and then selected theories that may similarly affect machine ``cognition''. These theories serve as the basis for us to hypothesize on how aesthetic elements influence model behavior.
Following the in-depth observation approach of the BERTology field \cite{rogers2021primer, htut2019attention, ettinger2020bert, jin2020bert, sun2020adv} towards the BERT model \cite{devlin2019bert}, we examine LayoutLMv3 \cite{huang2022layoutlmv3}, the current state-of-the-art document AI model, to test our hypotheses. Although focusing on a single model does not confirm the generalizability of insights, under resource constraints, it allows us to inspect the details of model behavior, laying a deeper foundation for future comparative analysis.
In summary, our primary contributions include: 
\begin{enumerate}
  \item A compilation of relevant research and hypothesis generation on how document design affects document understanding, with a focus on \emph{legibility} and \emph{layout quality};
  \item The curation and development of quantitative measures related to document aesthetics;
  \item An exploratory model analysis evaluating the correlation between document aesthetics and model confidence.
\end{enumerate}
The results indicate that while text size contrast impacts human attention, this is not significantly mirrored in the models tested. Furthermore, it is shown that incorporating image modality proves beneficial, particularly in classifying documents with poor alignment. More findings are discussed in Section \ref{sec:results}.
However, it is crucial to clarify that this paper does not seek to provide definitive conclusions, but rather to draw attention to model analysis grounded in document design theories, inspiring future research directions.

\section{Background}
In this chapter, we review document design principles and their influence on human perception. From this foundation, we introduce hypotheses that probe the impact of design aesthetics on model confidence, establishing the basis for our exploratory model analysis.

\subsection{Document Design: Theory, Criteria, and Hypotheses}
\label{chap:design-theory}
Document design can be conceptualized into three structures \cite{peels1985document, mao2003document}: 
(1) \emph{natural structure}: the expression of ideas in natural language, i.e., text; (2) \emph{logical structure}: hierarchical relationships between text segments, such as titles, chapters, sections, and paragraphs; (3) \emph{physical structure}: the visual presentation of content, such as fonts, images, and layout.
To analyze how multimodal models respond to document aesthetics, we focus on the physical structure of document design. Of the various criteria in the field, \emph{legibility} and \emph{layout quality} are the most prominently researched and have numerous established quantitative measures for their assessment.

Legibility refers to the visual clarity of the document and the ease of distinguishing individual characters in text. Choosing the appropriate font size and style allows readers to comprehend the text more rapidly \cite{tinker1963legibility}. Generally, larger fonts and uppercase characters offer higher legibility by making letters more visually distinct. \cite{bernard2001effects, sheedy2005text}. However, human reading studies often focus on determining the optimal font selection for reading entire body, and it is commonly believed that using lowercase characters \cite{babayigit2019reading} sized 10-12 enhances reading speed \cite{chandler2001running}. When applied to model behavior analysis, the concept of legibility naturally directs us to the accuracy of Optical Character Recognition (OCR) and its impact on model comprehension. Practically, prediction errors often stem from image noise. Given that document AI systems frequently encounter low-quality scanned document images, it becomes imperative to assess and explain their robustness to image noise \cite{jaume2019funsd}. 
In light of this, we propose and aim to test our hypothesis: 

\begin{itemize}
    \item \textbf{H1:} \emph{The degradation in the quality of document images negatively impacts legibility and the model's confidence.} 
\end{itemize}

Font-size contrast also affects legibility. 
For example, in a warning passage, when the font size difference between the signal word and the body is too large, readers may tend to overlook the body message \cite{braun1992likelihood}. This is relevant to machine reading behavior, 
as it is linked to the model's scale-awareness, 
an essential topic in computer vision \cite{lindeberg2013scale}. Also, larger text may dominate readers' attention, and this could likewise mislead the models' attention mechanism. 
Hence, we present the following hypothesis: 

\begin{itemize}
    \item \textbf{H2:} \emph{Font-size contrast overuse disrupts the model's attention, reducing its confidence.} 
\end{itemize}

The layout quality, on the other hand, pertains to the overall organization of the document, including the arrangement of text blocks, headings, images, and other visual elements. Akin to legibility, the fundamental research aspect of this criterion involves suitable selection of line length, line spacing, and so forth \cite{tinker1967bases, hartley2013designing}. 
Discussions on line length further extend to the comparison between single-column and multi-column layouts \cite{foster1970study, baker2005multiple}. 
Another vital aspect lies in its role in aiding content navigation. 
An intuitive spatial representation of the elements enables readers to index the information, thereby enhancing reading performance \cite{kennedy1992spatial, KENNEDY2003193}. 
The relative positioning of the elements should adhere to \emph{the principle of contiguity}, guiding the reading order that supports the logical structure of the document \cite{mayer2005principles}. 
Techniques such as \emph{grouping} and \emph{alignment} can be applied, in accordance with Gestalt psychology principles \cite{bruce2003visual}, to improve cohesion and predictability of the content \cite{dondis1974primer} by strategically arranging similar elements in close spatial proximity. 
Authors are encouraged to follow layout conventions \cite{wright1999psychology} to avoid extra cognitive load as it produces unexpected structures \cite{dillon2002designing}, 
and avoid overly complex layouts, which can hinder efficient prediction of content \cite{bonsiepe1968method, tullis1984predicting}.
Complementing legibility, our analysis from a perspective of layout quality emphasizes principles that promote content navigation. 
Based on the above, we formulate testable hypotheses:

\begin{itemize}
    \item \textbf{H3:} \emph{Misalignment impedes content comprehension and the model's prediction confidence.} 
    \item \textbf{H4:} \emph{As layout complexity escalates, the model's confidence in predictions decreases.} 
\end{itemize}

\subsection{Document AI}
Document AI \cite{cui2021document} (also known as Document Intelligence) systems are designed to automate tasks related to document processing and understanding. In business and societal applications, the documents of interest often contain not only text but also useful visual information. Such tasks are also known as Visually-rich Document Understanding (VrDU), with topics including Key Information Extraction (KIE) \cite{park2019cord, jaume2019funsd, huang2019icdar2019}, Document Layout Analysis (DLA) \cite{li2020docbank, zhong2019publaynet, pfitzmann2022doclaynet}, and Document Visual Question
Answering (DocVQA) \cite{mathew2021docvqa}. 
Recently, the state-of-the-art technologies are increasingly leveraging Transformer-based multimodal architectures to concurrently integrate text, layout, and image inputs for building pre-trained models, such as LayoutLMs \cite{xu2021layoutlmv2,huang2022layoutlmv3} and Docformer \cite{appalaraju2021docformer}. However, it is challenging to obtain human-centric interpretations about the inner workings of these neural networks. 

The current behavior analysis of document AI systems mainly relies on ablation studies to assess the accuracy impact of individual modalities. Differences in performance can be observed by altering various parameters in isolation, such as pre-training tasks, feature types, or other architectural elements, e.g., embeddings types, attention layers, or relation heads/scorers \cite{huang2022layoutlmv3, Gemelli_2023, appalaraju2021docformer, zhang2021entity, luo2023geolayoutlm, peng2022ernielayout}, thus confirming the effectiveness of different modalities. 
Despite the aforementioned studies, many behavior analyses of text-only pre-trained models have not yet been applied to document AI models, such as self-attention observation \cite{htut2019attention, kobayashi2020attention}, reliability analysis \cite{jin2020bert, sun2020adv}, and linguistic knowledge probing \cite{lin2019open, belinkov2022probing}. Note that, as we shift our focus towards VrDU, incorporating document aesthetics into the analyses becomes essential. This integration is key to bridging the gap in how models interpret visual elements, aligning it more closely with human cognitive processes. Such an approach is vital for the development of VrDU models that not only mirror human perception and interpretative behaviors but also significantly enhance user interaction with document analysis tools driven by these models.

\section{Exploratory Model Analysis}
To illuminate the role of document aesthetics in understanding model behavior, we demonstrate an exploratory analysis with the aim to test our hypotheses and gain insights. Given its suitability across all the hypotheses, we selected correlational analysis as our primary method. 
Specifically, we collected existing measures to quantify aesthetic factors and assessed their \textbf{correlation with model prediction confidence}. However, most of these are document-level measures, which average out effects across the entire page and might obscure local phenomena. Therefore, when a hypothesis test demands more granular information, we adapted concepts from literature to develop element-level measures.
Further details are provided below.

\subsection{Datasets and Settings}
\subsubsection{Datasets}
As the hypotheses may exhibit more relevancy to certain tasks, we conducted our analyses on two VrDU datasets, each corresponding to a different classification task. To obtain text data along with bounding boxes, we performed OCR on document images using ABBYY FineReader.\footnote{A commercial OCR software. \url{https://pdf.abbyy.com/}} The summaries of the datasets are listed below: 

\begin{enumerate}
  \item \textbf{FUNSD} \cite{jaume2019funsd}, a KIE dataset, consisting of 199 (149/50 for training/test; 30 samples from the training set are used for validation) noisy, scanned forms and a total of 9,707 annotated semantic entities. This dataset is widely used as a benchmark dataset for tasks such as OCR, spatial layout analysis, and entity extraction/linking. We have focused on the task of \textbf{entity labeling}, i.e., grouping words into semantic entities and labeling them as \textit{header}, \textit{question}, \textit{answer}, or \textit{other}. The statistics of entity labels are presented in Table \ref{tab:data-funsd}. To incorporate a more realistic scenario and assess the impacts of document aesthetics, we opted to derive text and coordinates using ABBYY FineReader, thereby accounting for OCR errors, instead of relying on the high-quality input text and coordinates provided with the dataset.
  \item \textbf{IDL},\footnote{https://www.industrydocuments.ucsf.edu/} a vast collection of documents created by industries which influence public health. It has fostered multiple datasets for VrDU tasks, such as RVL-CDIP \cite{harley2015evaluation} and DocVQA \cite{mathew2021docvqa}. We have focused on the task of \textbf{document page classification}, i.e., labeling pages originating from documents with a single class, using a subset of about 15K (80\%/10\%/10\% for training/validation/test) OCR'd documents from OCR-IDL \cite{biten2023ocr}. Our dataset emulates RVL-CDIP, which contains 16 document categories. We sampled single-category documents from IDL that are labeled as one of these categories. The label statistics are displayed in Table \ref{tab:data-idl}.
\end{enumerate}

\begin{table}[t!]
  \caption{FUNSD: Class distribution of the semantic entities. The dataset has two splits, i.e., training and test, and four classes of semantic entities, i.e., header, question, answer, and other.}
  \label{tab:data-funsd}
  \begin{center}
  \begin{tabular}{cccccc}
      \toprule
      \textbf{Split} & \textbf{Header} & \textbf{Question} & \textbf{Answer} & \textbf{Other} & \textbf{Total} \\ 
      \midrule
      \multirow{1}{*}{Training}    
          & 441 & 3,266 & 2,802 & 902 & 7,411 \\ 
      \multirow{1}{*}{Test}
          & 122 & 1,077 & 821 & 312 & 2,332  \\ 
      \bottomrule
  \end{tabular}
  \end{center}
\end{table}

\begin{table}[t!]
  \caption{IDL: Class distribution of page types. The dataset has three splits, i.e., training, validation, and test, and 16 classes of page types, e.g., news article, invoice, and resume.}
  \label{tab:data-idl}
  \begin{center}
  \begin{tabular}{cccc}
      \toprule
      \textbf{Category} & \textbf{Training} & \textbf{Validation} & \textbf{Test} \\ 
      \midrule
      News Article & 877 & 110 & 110 \\ 
      Memo & 874 & 109 & 109 \\ 
      Advertisement & 869 & 109 & 108 \\
      Form & 833 & 104 & 104 \\
      Letter & 829 & 104 & 103 \\
      File Folder & 824 & 103 & 103 \\
      Email & 810 & 101 & 102 \\
      Scientific Report & 785 & 98 & 98 \\
      Specification & 778 & 97 & 97 \\
      Questionnaire & 754 & 95 & 94 \\
      Budget & 734 & 92 & 92 \\
      Invoice & 715 & 89 & 90 \\
      Presentation & 657 & 82 & 82 \\
      Handwritten & 591 & 74 & 74 \\
      Scientific Publication & 534 & 67 & 67 \\
      Resume & 389 & 48 & 49 \\
      \midrule
      \textbf{Total} & 11,853 & 1,482 & 1,482 \\
      \bottomrule
  \end{tabular}
  \end{center}
\end{table}

\subsubsection{Model Configurations}
We chose LayoutLMv3 \cite{huang2022layoutlmv3}, the current state of the art, to test our hypotheses. 
An ablation study was conducted by masking modalities during fine-tuning. 
Specifically, we fine-tuned the pre-trained checkpoint on Hugging Face\footnote{LayoutLMv3-base:  \url{https://huggingface.co/microsoft/layoutlmv3-base}} using the Trainer API\footnote{https://huggingface.co/docs/transformers/main\_classes/trainer} and masked one modality at a time. 
The image modality is masked by replacing the input with a black image, and the layout modality is masked by zeroing out the bounding box coordinates for all input tokens, 
to nullify their effects.

Using this, the model is fine-tuned in four settings of modality combinations, i.e., \textbf{T+L+I}, \textbf{T+L}, \textbf{T+I}, and \textbf{T}, where T/L/I stands for Text/Layout/Image, respectively. We fine-tuned the model once for each setting, from which the results are reported. The hyperparameters of the model are detailed in Appendix \ref{appx:model-conf}. The outcomes of fine-tuning in terms of classification performance metrics, namely precision, recall, and F1-score, are shown in Appendix \ref{appx:model-perf}.

\subsection{Measures}
\label{sec:measure}
\subsubsection{Image Noise}
\label{sec:measure-noise}
In scenarios without reference images for quality comparison, No-Reference Image Quality Assessment techniques \cite{kamble2015no} provide a practical solution. These methods assess images for specific types of distortions directly from the data. A common sign of image noise in documents is the presence of high-frequency components, for example, rapid changes in pixel intensity indicative of edges and texture. When an image exhibits abundant abrupt changes that do not align with its underlying structure, this is identified as noise \cite{tan2019image}.

In the absence of reference images, and considering that noise in our dataset is predominantly marked by numerous high-frequency components rather than other forms of noise, such as blurriness, we focus on analyzing high-frequency components to assess noise levels. We employ the 2D Discrete Fourier Transform (DFT) method \cite{de2013image} to convert spatial data into the frequency domain. The DFT is computed using the Fast Fourier Transform \cite{2014opencv}, which facilitates the identification of high-frequency components. A higher measure value denotes a greater presence of high frequency noise in the image.

\subsubsection{Font Size and Contrast}
\label{sec:measure-contrast}
We derive font size data from ABBYY FineReader's \emph{fs} attribute for each line. 
Following \citet{braun1992likelihood}, we test the distraction effect. As there is no well-known measure tailored for this, we developed a measure that compares the sizes of two nearest elements, defined as $T_i = (S_n - S_i)/S_i$, where $S_i$ is the size of element $i$ and $S_n$ denotes the size of its nearest neighbor. 
 
\subsubsection{Misalignment}
We first adhere to research conventions by calculating an alignment score and then take its complement to determine the misalignment score $M = 1 - Alignment(\cdot)$.
For IDL, we utilize the alignment measure proposed by \citet{ngo2003modelling} for popularity over others \cite{harrington2004aesthetic, balinsky2009aesthetic}. 
They introduce a formula where the alignment score is $1$ for single-element documents; otherwise, it is $1 - (n_{\text{vap}} + n_{\text{hap}}) / 2n$, with $n_{\text{vap}}$ and $n_{\text{hap}}$ as the counts of vertical and horizontal alignment points, respectively, and $n$ as the number of elements.
A lower count of alignment points corresponds to a higher alignment score, indicating better document regularity.

However, in response to the need for testing H3 at the element level in FUNSD, we introduced the Element-level Alignment Measure algorithm, an adaptation of the measure mentioned above. This approach shifts the assessment from a document-wide perspective to an examination of individual elements, thereby facilitating a detailed element-level hypothesis testing. The algorithm operates through two primary functions: \texttt{ALIGN} and \texttt{MEASURE}. 
The \texttt{ALIGN} function takes as input a list of bounding boxes representing document elements, a specified alignment mode (top-left, center, or top-right) to determine the reference point used for alignment, and a tolerance threshold for alignment deviation. Then, it iterates through these reference points to establish anchor points. These serve as the basis for aligning elements, indirectly forming alignment groups based on shared anchor points. If a reference point's nearest anchor point is within the tolerance threshold, it is considered aligned; otherwise, it becomes a new anchor point. The function returns the anchor points of the input elements, setting the stage for a more detailed analysis of alignment groups in the next step. 
Following this, the \texttt{MEASURE} function computes the alignment score for each element. For each alignment mode, it first calls the \texttt{ALIGN} function to obtain the set of anchor points. It then computes the alignment score for each element based on its membership in these groups, assessing the proportion of boxes aligned with the same anchor point. The final score for each element is the maximum score across all modes, offering an understanding of the element's optimal alignment scenario through its association with the most cohesive alignment group. The pseudocode is shown in Algorithm \ref{alg:alignment}.

\begin{algorithm}[h!]
  \caption{Element-level Alignment Measure}\label{alg:alignment}
  \begin{algorithmic}[1]
  \State $\textbf{function}\ \texttt{ALIGN}(boxes,mode,tolerence)$
  \State \quad $\#\ \text{assumes}\ boxes\ \text{are ordered from top to bottom;}$ 
  \State \quad $\#\ mode\ \text{is either}\ \textsc{topleft},\ \textsc{center},\ \text{or}\ \textsc{topright}.$
  \State \quad $refPoints \gets [box.getRefPoint(mode)\ \textbf{for}\ box\ \textbf{in}\ boxes]$
  \State \quad $anchors \gets List()$
  \State \quad $\textbf{for}\ pt\ \textbf{in}\ refPoints:$
  \State \quad \quad $anchor,distance \gets pt.getNearestPoint(anchors)$
  \State \quad \quad $\textbf{if}\ distance > tolerence:\ anchor \gets pt$
  \State \quad \quad $anchors.append(anchor)$
  \State \quad $\textbf{return}\ anchors$
  \State
  \State $\textbf{function}\ \texttt{MEASURE}(boxes,tolerence)$
  \State \quad $scores \gets [0\ \textbf{for}\ box\ \textbf{in}\ boxes]$
  \State \quad $\textbf{for}\ mode\ \textbf{in}\ [\textsc{topright},\textsc{center},\textsc{topleft}]:$
  \State \quad \quad $anchors \gets \texttt{ALIGN}(boxes,mode,tolerance)$
  \State \quad \quad $\textbf{for}\  i\ \textbf{in}\ range(boxes.length):$
  \State \quad \quad \quad $score \gets anchors.count(anchors[i])/boxes.length$
  \State \quad \quad \quad $\textbf{if}\ score > scores[i]: scores[i] \gets score$
  \State \quad $return\ scores$
  \end{algorithmic}
\end{algorithm}

Essentially, the Element-level Alignment Measure algorithm calculates the proportion of an alignment group in a document. For instance, in the left of Figure \ref{fig:funsd}, 5 of 24 left-aligned elements (in red) form a group, yielding a score of $5/24=0.208$; the complement $1-0.208=0.792$ denotes the misalignment score.

\subsubsection{Layout Complexity}

We compared two complexity measures proposed by \citet{ngo2003modelling} and \citet{bonsiepe1968method}, choosing the latter for its nonparametric nature. The idea is to classify elements based on common heights, widths, and distances to document edges (i.e., their x and y positions, reflecting horizontal and vertical distances, respectively), then calculate a modified version of Shannon's entropy from the resulting distribution, defined as $\Omega = -N \sum_{i=1}^{i=n} p_i \cdot log_2 p_i$, where $N$ is total number of elements; $n$ is total number of classes; $p_i$ denotes the frequency of $i$-th class. 
For example, consider a document with two elements whose corresponding bounding boxes are (10, 10, 5, 2) and (10, 14, 8, 2), denoted by ($x$, $y$, $width$, $height$). There are 1/2/2/1 unique x-position/y-position/width/height, respectively. Using these, we build the classes and compute the complexity with respect to each aspect. For instance, the y-position complexity is calculated as $-2 * \left( 0.5 \cdot log_2(0.5) + 0.5 \cdot log_2(0.5) \right)$, and the overall complexity is the sum of the complexity scores for all four aspects. 

\subsubsection{Model Confidence}
To measure model confidence, we applied normalized entropy, which is commonly used for quantifying model uncertainty \cite{abdar2021review}. Specifically, our measure is defined as $C = 1 - NormalizedEntropy(P)$, where $P$ is the prediction distribution across classes. The higher the uncertainty, the lower the confidence. 

\begin{figure}[t!]
  \centering
  \includegraphics[width=0.95\linewidth]{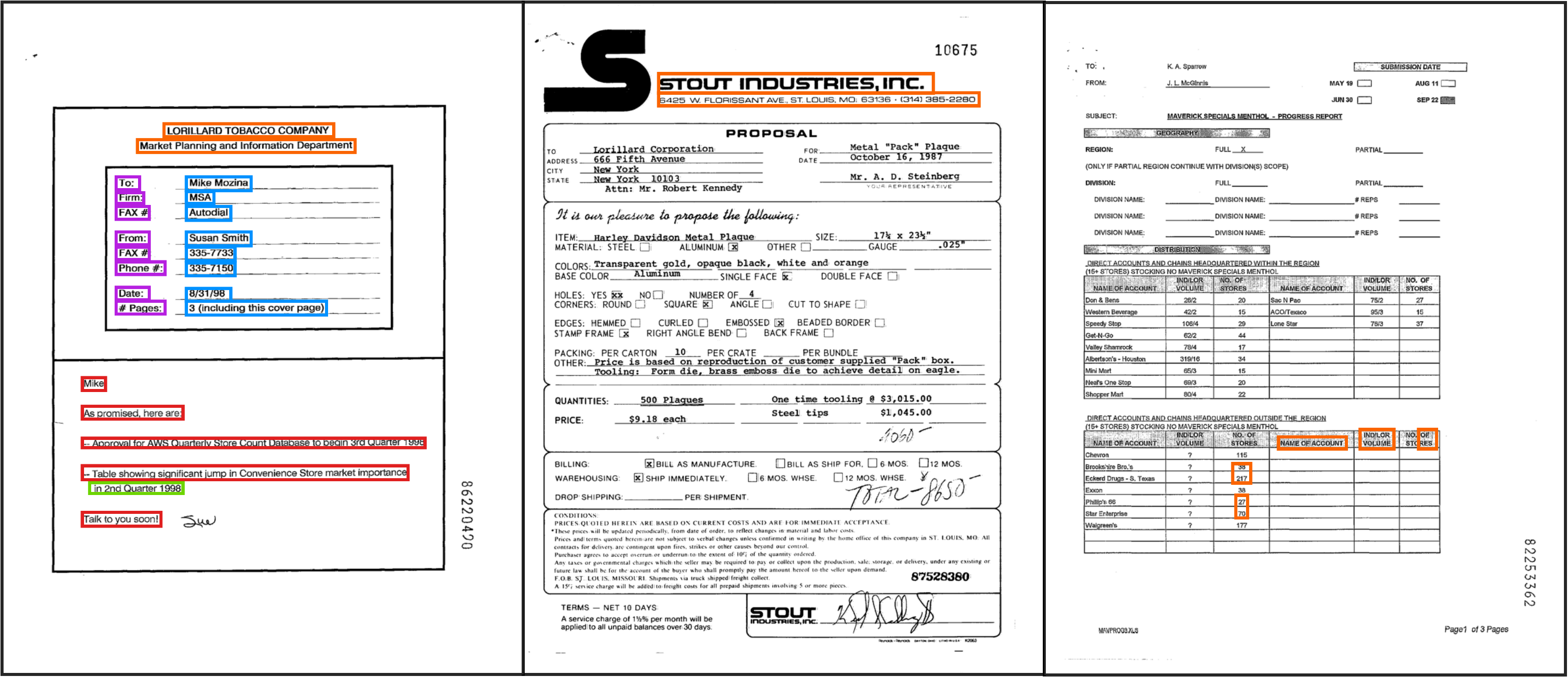}
  \caption{FUNSD: Examples. On the left, line-level elements and their alignment groups, identified by Algorithm \ref{alg:alignment}, are marked in different colors; elements in the orange group at the top align by their center reference points, while elements in the remaining groups align by their left reference points. In the middle, it is a case where an element exhibits excessive contrast, marked in orange boxes. The nearest element of the address line is the company name, emphasized with a significantly larger font size, resulting in a high contrast score as calculated by the formula in Section \ref{sec:measure-contrast}. On the right, a case of high contrast due to OCR errors is depicted, where two lines can be mistakenly recognized as one element, such as ``IND/LOR'' combined with ``VOLUME'', highlighted in orange. This demonstrates real-world instances where OCR inaccuracies lead to confusion in models' understanding of layout.}
  \label{fig:funsd}
\end{figure}

\begin{figure}[t!]
  \centering
  \includegraphics[width=0.95\linewidth]{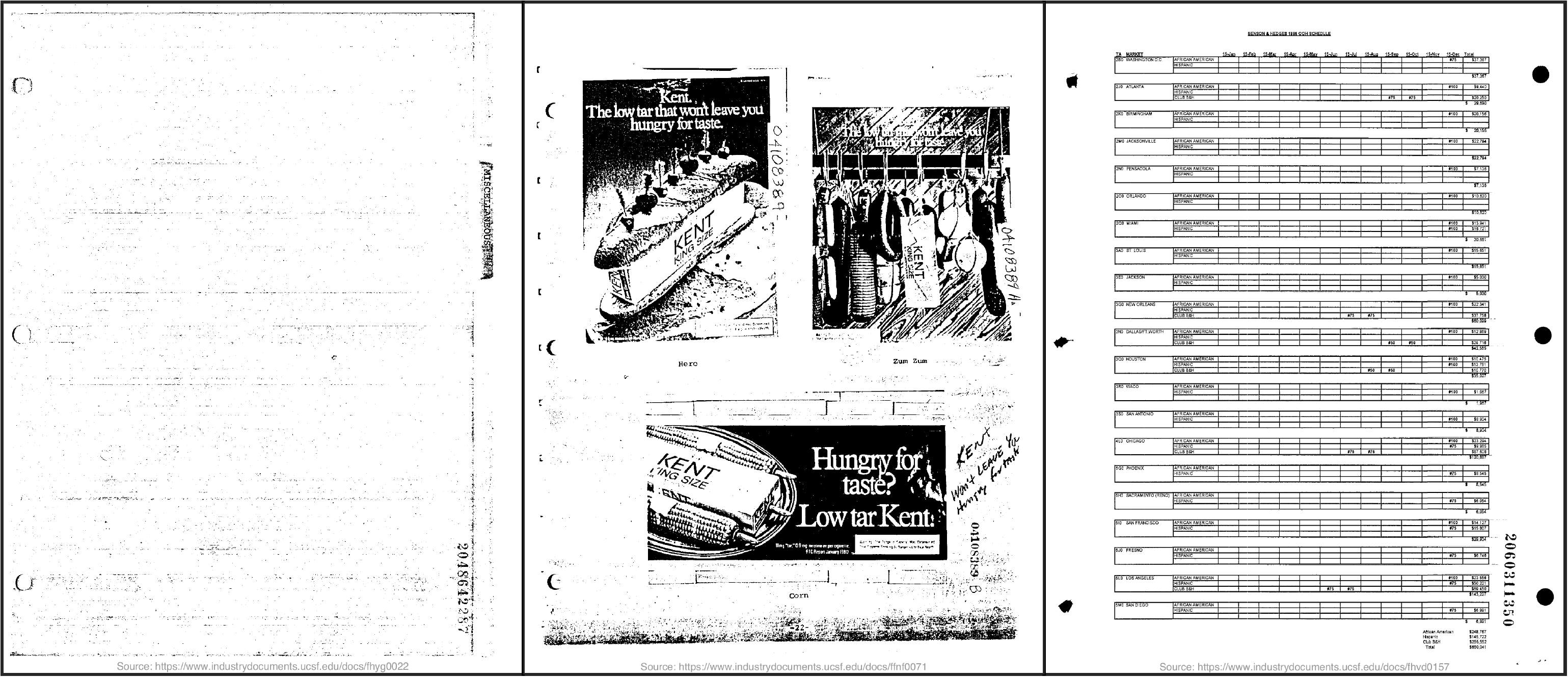}
  \caption{IDL: Examples. One the left, it is an instance of ``white'' file folders, which in general have lower noise scores as measured by the method described in Section \ref{sec:measure-noise}. This lower score is attributed to their large areas of uniform color (i.e., white pixels) and minimal abrupt changes. In the middle, an advertisement document is shown, characterized by rich graphic elements and irregular text arrangement, which typically results in higher noise scores. On the right, On the right, a form with high complexity is displayed, assessed by Bonsiepe's formula \cite{bonsiepe1968method}, but it also exhibits a higher quality of alignment as evaluated by Algorithm \ref{alg:alignment}.}
  \label{fig:idl}
\end{figure}

\section{Results and Case Study}
\label{sec:results}
Due to the non-normal and non-linear relationships between the aesthetic measures and the model's confidence, instead of Pearson correlation coefficient, we opt for \textbf{Spearman's $\rho$, and the p-values were tested}.
Outliers are removed to reduce the overwhelming impact arising from factors such as OCR errors (e.g., misrecognized or missing text blocks) and data inaccuracies (as reported by \citet{larson2023evaluation}, potentially 8.1\% mislabeled samples in RVL-CDIP, likely extending to IDL), which introduce extreme values in document aesthetic measures and anomalously low prediction confidences. This approach ensures that the analysis focuses on the core data trends.
The results of the correlation analysis are presented in Table \ref{tab:corr}. We discuss the details below:

\subsection{Hypothesis 1: Image Noise}
Since the major issue related to image noise in the KIE task (i.e., FUNSD) is trivially OCR errors, which have a dominant impact on model performance, we pivot towards the document page classification task (i.e., IDL) to test the noise effect, focusing on identifying beneficial modalities to address these issues.
We applied the document-level measure of image noise as in Section \ref{sec:measure-noise} and the results show that 
the image modality (included in \textbf{T+L+I} and \textbf{T+I}) displayed sensitivity to noise, while in its absence (i.e., \textbf{T+L} and \textbf{T}) the correlation was insignificant. 
Upon closer comparison, we observed the image modality excelling especially in low-noise scenarios (e.g., file folders; the left in Figure \ref{fig:idl}), where other modalities had limited text blocks to extract features. 

\subsection{Hypothesis 2: Font-size Contrast}
Given its relevance to the KIE task, requiring the model to understand the context in neighboring content, we test the distraction effect on FUNSD.
Using the proposed measure described in Section \ref{sec:measure-contrast}, we compute the contrast score for each element in a document and evaluate their correlation with element-level confidence of predictions. 
It is important to note that moderate contrast can aid comprehension, and only extreme contrast may distract (i.e., when it is ``overused'', as stated in \textbf{H2}). Therefore, we assessed the correlation at various levels of excessiveness, gauged by standard deviation (Stdev), and only anticipate the hypothesized effect when contrast exceeds a certain threshold, e.g., 1 Stdev. 
We report the results for 3 Stdev, as no notable correlation exists below this level, and observed that the confidence of \textbf{T+L+I} model exhibits sensitivity to font-size contrast.
After further scrutiny, we discovered that the high contrast cases often occurred near stylish \emph{headers} (e.g., the middle in Figure \ref{fig:funsd}), which seemingly echo the neglect effect \cite{braun1992likelihood}, and were sometimes induced by OCR errors, which mistakenly merged adjacent multi-line elements into one nonsensical, larger font size line (e.g., the right in Figure \ref{fig:funsd}). 
Although our analysis aims to suggest appropriate contrast levels, the findings from extreme cases with a small size of samples (i.e., the 3 Stdev threshold excludes most samples) conclude only that the model is minimally impacted by font size contrast.

\begin{table}[t!]
    \caption{Spearman Correlation Analysis between Aesthetic Measures and Model Confidence: the hypotheses of image noise (\textbf{H1}), font-size contrast (\textbf{H2}), misalignment (\textbf{H3}), and layout complexity (\textbf{H4}) are tested; \textbf{bold} values denote significance (p-value < 0.05). T/L/I stands for Text/Layout/Image, respectively, indicating the modalities that are used for model input.}
    \label{tab:corr}
    \begin{center}
    \begin{tabular}{cccccc}
        \toprule
        \textbf{Hypo.} & \textbf{Dataset} & \textbf{T+L+I} & \textbf{T+L} & \textbf{T+I} & \textbf{T} \\ 
        \midrule
        \multirow{1}{*}{\textbf{H1}}   
            & IDL & \textbf{-0.19} & -0.02 & \textbf{-0.14} & +0.03 \\ 
        \midrule
        \multirow{1}{*}{\textbf{H2}}
            & FUNSD & \textbf{-0.44} & -0.01 & -0.15 & -0.20  \\ 
        \midrule
        \multirow{2}{*}{\textbf{H3}} 
            & FUNSD & \textbf{-0.16} & \textbf{-0.11} & -0.00 & -0.02   \\
            & IDL & +0.03 & -0.04 & \textbf{+0.11} & \textbf{-0.16}   \\
        \midrule
        \multirow{1}{*}{\textbf{H4}} 
            & IDL & \textbf{-0.17} & \textbf{-0.08} & \textbf{-0.39} & +0.04   \\
        \bottomrule
    \end{tabular}
    \end{center}
\end{table}

\subsection{Hypothesis 3: Misalignment}
We use Algorithm \ref{alg:alignment} to closely examine the impact of misaligned elements on FUNSD.
The results showed that \textbf{T+L+I} and \textbf{T+L} confidence correlates negatively with misalignment scores, suggesting that the layout modality is pivotal in responding to alignment quality in the KIE task, whereas image modality alone had minimal effect. However, in page classification (IDL), poorly aligned documents only adversely affect the \textbf{T} model and, surprisingly, boost the \textbf{T+I} model's confidence. Upon closer inspection of the latter, we found that these instances were mostly advertisements (e.g., the middle of Figure \ref{fig:idl}), which typically have distinct visual patterns. 

\subsection{Hypothesis 4: Layout Complexity}
Lastly, we examined layout complexity
using Bonsiepe's formula. 
Since this aggregates the layout information across the whole page, we consider the page classification task more relevant 
and test \textbf{H4} on IDL.
As a result, we see significant impacts on all multimodal models, i.e., \textbf{T+L+I}, \textbf{T+L}, and \textbf{T+I}. This suggests that layout complexity can mirror the difficulty in differentiating documents in IDL. Nonetheless, the impact of complex content diminished when layout modality was included. Through case study, we observed that complex content can be well-aligned (e.g., the right in Figure \ref{fig:idl}), where the alignment patterns recognized through layout modality may mitigate the impact from complexity. 
Although analyzing model behavior on FUNSD under identical settings might not resonate with the KIE task's emphasis on local context for aiding comprehension, we ventured an attempt and, as anticipated, found no significant correlation with model confidence.

\section{Conclusion and Future Work}
In this study, we examined model behavior through the lens of document design theories, focusing on legibility and layout quality for a more user-oriented understanding of the models. Our research provides statistical evidence of a correlation between the above aspects and model predictions. We observed that different modalities react variably to these aspects, with one modality's compensatory effect sometimes balancing another's impact. For example, layout information becomes vital in complex layouts, while the significance of other modalities may decrease. Results reveal other intriguing insights, e.g., large fonts can induce distraction, mirroring human perception \cite{braun1992likelihood}. While the results are statistically validated, this study is exploratory in nature, offering many opportunities for further research.

Future work could delve into other elements of document design, including color, boldness, tables, graphs, readability \cite{dubay2004principles}, and more. However, ideal metrics for these aspects that resonate with document design principles are not yet established and require comprehensive study. We also identified a notable gap that needs filling in benchmark datasets, potentially attributable to insufficient attention to this topic and the challenges from its interdisciplinary nature. To broaden our findings, we plan to conduct in-depth analyses, such as explanatory analysis, and test additional VrDU models, such as UDOP \cite{tang2023unifying} and Pix2Struct \cite{lee2023pix2struct}. In addition, to uphold scientific rigor, it is also crucial to validate the measures we developed in this study to ensure their consistency with human judgment. Our objective is to inspire the creation of document AI models that align with human cognitive processes, which can markedly enhance the overall user experience with document analysis tools and foster a wide range of applications.

\section{Limitations}
Our experiments were carried out using a single VrDU model, namely LayoutLMv3. Therefore, our findings may not generalize to other models that have different architectures or pre-training approaches. To extend these findings to a wider range of document AI models, we acknowledge the potential need for modifications in the masking strategy. These modifications should align with variations in pre-training tasks and the implementation of different modalities. For example, if a model does not represent the layout modality through bounding boxes, as is the case with LayoutLMv3, adjustments may be necessary. Additionally, our study is limited by elements of document design that present challenges for quantitative measurement. There may also be other significant factors that warrant inclusion in this study but cannot be incorporated due to the lack of appropriate measurement techniques in the existing literature.

\newpage
\bibliography{references}

\newpage
\appendix

\section{Hyperparameter Configurations}
\label{appx:model-conf}

This section provides an overview of the training configurations employed for fine-tuning the LayoutLMv3 model (\textit{layoutlmv3-base}) on the two datasets, i.e., FUNSD and IDL, which was performed using Hugging Face Trainer API. For all models trained in the ablation study, Table \ref{tab:hyperparams} exhibits the ranges of key hyperparameter values. This information serves as a guide for replicating the training process.

\begin{table}[ht!]
  \caption{Hyperparameter configurations for LayoutLMv3 fine-tuning on FUNSD and IDL datasets.}
  \label{tab:hyperparams}
  \centering
  \begin{tabular}{lcc}
      \toprule
      \textbf{Hyperparameter} & \textbf{FUNSD} & \textbf{IDL} \\ 
      \midrule
      Epochs & [6,16] & [8,10] \\
      Batch Size & 2 & 10 \\
      Gradient Accumulation Steps & 1 & 1\\
      Learning Rate & [1e-4,1e-5] & 2e-5 \\
      Optimizer & AdamW & AdamW \\
      FP16 & True & True \\
      \bottomrule
  \end{tabular}
\end{table}

\section{Model Performance: Precision, Recall, and F1-Score}
\label{appx:model-perf}

The primary aim of our research was not to inspect or enhance model performance metrics, but rather to analyze the correlations between prediction confidence and aesthetic measures. Nevertheless, for completeness, we present the macro-averaged precision, recall, and F1 scores of our fine-tuned models for each task, as evaluated on the test split of the corresponding dataset. Table \ref{tab:model-perf} presents the details.

\begin{table}[h!]
  \caption{Macro-averaged precision, recall, and F1-score of each fine-tuned model across all classes on FUNSD and IDL datasets.}
  \label{tab:model-perf}
  \centering
  \begin{tabular}{c|ccc|ccc}
      \toprule
        & \multicolumn{3}{c}{\textbf{FUNSD}} & \multicolumn{3}{c}{\textbf{IDL}} \\
      \textbf{Model} & \textbf{Precision} & \textbf{Recall} & \textbf{F1-score} & \textbf{Precision} & \textbf{Recall} & \textbf{F1-score} \\ 
      \midrule
      \textbf{T+L+I} & 0.834 & 0.835 & 0.834 & 0.923 & 0.920 & 0.921 \\
      \textbf{T+L} & 0.820 & 0.821 & 0.821 & 0.908 & 0.907 & 0.907 \\
      \textbf{T+I} & 0.770 & 0.769 & 0.769 & 0.916 & 0.914 & 0.915 \\
      \textbf{T} & 0.762 & 0.764 & 0.762 & 0.879 & 0.874 & 0.875 \\
      \bottomrule
  \end{tabular}
\end{table}

\end{document}